\documentclass[sn-mathphys-ay,table,xcdraw]{sn-jnl}

\usepackage{graphicx}%
\usepackage{multirow}%
\usepackage{amsmath,amssymb,amsfonts}%
\usepackage{amsthm}%
\usepackage{mathrsfs}%
\usepackage[title]{appendix}%
\usepackage{xcolor}%
\usepackage{textcomp}%
\usepackage{manyfoot}%
\usepackage{booktabs}%
\usepackage{lscape}
\usepackage[caption=false,font=normalsize,labelfont=sf,textfont=sf]{subfig}

\usepackage{bm}
\usepackage{enumitem}
\usepackage[normalem]{ulem}
\useunder{\uline}{\ul}{}
\usepackage{pifont}
\newcommand{\cmark}{\ding{51}}%
\newcommand{\xmark}{\ding{55}}%
\newcommand{\mr}[1]{\textcolor{black}{#1}}
\newcommand{\mrr}[1]{\textcolor{black}{#1}}

\begin{document}

\title[Article Title]{Sports-QA: A Large-Scale Video Question Answering Benchmark for Complex and Professional Sports}


\author[1]{\fnm{Haopeng} \sur{Li}}

\author[2]{\fnm{Andong} \sur{Deng}}

\author*[3]{\fnm{Jun} \sur{Liu}}

\author[3]{\fnm{Hossein} \sur{Rahmani}}

\author[4]{\fnm{Yulan} \sur{Guo}}

\author[5]{\fnm{Bernt} \sur{Schiele}}

\author[6]{\fnm{Mohammed} \sur{Bennamoun}}
 
\author*[7]{\fnm{Qiuhong} \sur{Ke}}\email{qiuhong.ke@monash.edu}

\affil[1]{\orgdiv{School of Computing and Information Systems}, \orgname{University of Melbourne}}

\affil[2]{\orgdiv{Center for Research in Computer Vision}, \orgname{University of Central Florida}}

\affil[3]{\orgdiv{School of Computing and Communications}, \orgname{Lancaster University}}

\affil[4]{\orgdiv{School of Electronics and Communication Engineering}, \orgname{Sun Yat-sen University}}

\affil[5]{\orgdiv{Department of Computer Vision and Machine Learning}, \orgname{Max Planck Institute for Informatics, Saarland Informatics Campus}}

\affil[6]{\orgdiv{School of Physics, Maths and Computing}, \orgname{University of Western Australia}}

\affil[7]{\orgdiv{Department of Data Science \& AI}, \orgname{Monash University}}


\abstract{Reasoning over sports videos for question answering is an important task with numerous applications, such as player training and information retrieval. However, this task has not been explored due to the lack of relevant datasets and the challenging nature it presents. Most datasets for video question answering (VideoQA) focus mainly on general and coarse-grained understanding of daily-life videos, which is not applicable to sports scenarios requiring professional action understanding and fine-grained motion analysis. In this paper, we introduce the first dataset, named Sports-QA, specifically designed for the sports VideoQA task. The Sports-QA dataset includes various types of questions, such as descriptions, chronologies, causalities, and counterfactual conditions, covering multiple sports. Furthermore, to address the characteristics of the sports VideoQA task, we propose a new Auto-Focus Transformer (AFT) capable of automatically focusing on particular scales of temporal information for question answering. We conduct extensive experiments on Sports-QA, including baseline studies and the evaluation of different methods. The results demonstrate that our AFT achieves state-of-the-art performance\footnote{Sports-QA is available at: \url{https://github.com/HopLee6/Sports-QA}}.}




\maketitle

\section{Introduction}


Sports video analysis has been attracting increasing attention in recent years
\citep{yuan2021spatio, li2021groupformer, koshkina2021contrastive, zhu2022fencenet, martin2020fine, wang2022shuttlenet,6516867}.
While research progress has been made on tasks such as
sports action recognition \citep{li2021multisports, zhu2022fencenet, rasmussen2022compressing}, reasoning over sports videos for question answering
has not been explored. As humans, we can not only recognize the actions of the players in sports videos but also understand the effects of the players' actions, explain why a team loses the score, and imagine what would happen under counterfactual situations. Our impressive capabilities in reasoning allow us to answer complex questions related to sports videos, which is crucial in applications such as obtaining crucial statistics in matches for player/team performance qualification, analyzing players' actions and team strategies for performance improvement, and efficiently retrieving information for audiences and analysts.
While sports video reasoning is clearly important, it is under-explored due to challenges and the lack of datasets.

\begin{figure}[tbp]
\centering
\includegraphics[width=0.9\columnwidth]{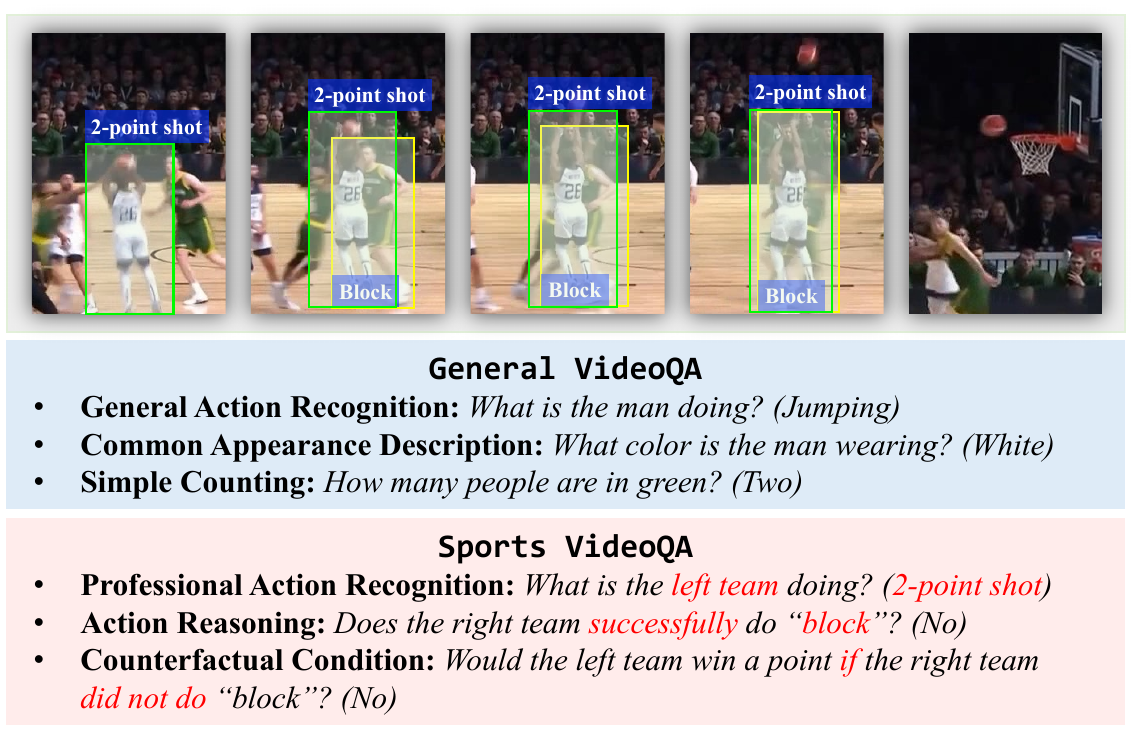}
\caption{
Illustrations of general VideoQA, which focus on common basic understanding, and sports VideoQA, which requires professional action understanding and action relation reasoning\protect\footnotemark. 
}
\label{qadiff}
\end{figure}

In this paper, we tackle sports VideoQA, focusing on reasoning over sports videos for question answering. This is distinct from general VideoQA, which centers on general video understanding, mainly encompassing descriptive and temporal aspects \citep{jang2017tgif, xu2017video, xiao2021next, yu2019activitynet}, as depicted in Fig. \ref{qadiff}. 
Compared to general VideoQA, sports VideoQA is more challenging because the questions may involve particular teams and professional actions, and it requires a fine-grained understanding of actions and intra/inter-team relation modeling. 
As pointed out in \citep{shao2020finegym}: 1) coarse-grained global video understanding is insufficient for sports analysis where the actions are rapid and hard to differentiate; 2) the models pre-trained on coarse-grained datasets are not always beneficial to fine-grained tasks. 
In the meantime, although significant strides have been made in the fine-grained recognition of individual actions \citep{shao2020finegym,liu2022fineaction,sun2017taichi}, these datasets may fall short in effectively evaluating the ability for temporal and causal reasoning over actions performed by individuals or teams in sports scenarios.
A large-scale dataset containing professional sports scenarios with detailed annotations of sports-relevant questions is thus required for explicit and comprehensive reasoning for sports VideoQA.

To address this gap, we introduce a novel dataset called Sports-QA, the first video QA dataset specifically focusing on sports activities. Leveraging sports videos and professional action labels from MultiSports \citep{li2021multisports} and FineGym \citep{shao2020finegym}, we construct our dataset. Both sources provide high-quality sports videos with well-defined action labels, allowing us to annotate essential attributes for each crucial action instance. Using these annotations, we generate QA pairs based on pre-defined templates covering aspects such as description, chronology, causality, and counterfactual conditions. The Sports-QA dataset is the first of its kind, comprising approximately 94K QA pairs, encompassing multiple sports and various question types for sports VideoQA. Table \ref{set} provides a comparison between Sports-QA and several widely-used VideoQA datasets.

\footnotetext[\thefootnote]{We treat the action ``\textit{block}" as a noun.}

\begin{table}[tbp]
\caption{\mrr{VideoQA datasets comparison. ``Auto/Man" represents ``automatic/manual". ``MC/OE" represents ``multi-choice/open-ended".}}
\label{set}
\begin{tabular}{@{}lcrrcc@{}}
\toprule
Dataset                           & Topic   & {\#Video} & {\#QA} & Annotation  & QA Task  \\ \midrule
MSRVTT-QA \citep{xu2017video}      & \multirow{4}{*}{General} & 10K                         & 244K                     & Auto        & OE       \\
MSVD-QA \citep{xu2017video}        &  & 2K                          & 51K                      & Auto        & OE       \\
NExT-QA \citep{xiao2021next}       &  & 5.4K                        & 52K                      & Man         & OE\&MC \\
TGIF-QA \citep{jang2017tgif}       &  & 72K                         & 165K                     & Auto\&Man & OE\&MC \\ \midrule
SportQA \citep{xia2024sportqa}& Sports& 0& 70K& Auto\&Man& MC       \\
SPORTU-text \citep{xia2024sportu}& Sports& 0& 0.9K& Man& MC       \\
SPORTU-video \citep{xia2024sportu}& Sports& 1.7K& 12K& Man& MC       \\
MovieQA \citep{tapaswi2016movieqa} & Movie   & 408                         & 15K                      & Man         & MC       \\
Social-IQ \citep{zadeh2019social}  & Social  & 1K                          & 8K                       & Man         & MC       \\
DramaQA \citep{choi2021dramaqa}    & Drama   & 24K                         & 16K                      & Auto\&Man & MC       \\
\textbf{Sports-QA (Ours)}                  & Sports  & 6K                          & 94K                      & Auto\&Man & OE       \\ \bottomrule
\end{tabular}
\end{table}

Compared with existing datasets, our Sports-QA provides new insights for VideoQA: 1) It encompasses questions related to specific terms and actions in both group activities (e.g., basketball) and single-person sports (e.g., gym). This requires models to possess the capability to capture dynamic patterns and perform reasoning in scenarios with varying numbers of interactions for question answering. 2) To achieve a comprehensive understanding of complex sports videos, Sports-QA includes diverse questions that involve video information at various temporal granularities. This encompasses global long-term temporal dependencies and fine-grained short-term ones. For instance, a question asking about the number of actions requires the model to capture global dependencies, while a question about the effect of a specific action relies on short-term temporal information. Furthermore, Sports-QA has the potential to be leveraged for other tasks. The pre-defined attributes of actions can be treated as multiple labels, allowing the dataset to be used for multi-label classification tasks for comprehensive action understanding. Additionally, based on the annotated action attributes, we can modify the templates from question answering to declarative sentences, generating descriptive or explanatory narrations and enabling comprehensive sports video captioning.

In addition to the dataset, we present a novel method for sports VideoQA. Recognizing that sports VideoQA demands the model to capture information from multiple frames at various temporal granularities, we introduce the Auto-Focus Transformer (AFT), featuring an Auto-Focus Attention mechanism. Conceptually, the model operates akin to a temporal magnifying glass with a dynamic focal length, allowing it to inspect the video to answer questions with diverse temporal dependencies. More specifically, our designed Auto-Focus Attention dynamically selects attention focus based on the question. This mechanism empowers the model to handle questions involving video information across various time spans.

The contributions are summarized as follows:

\begin{itemize}

    \item We contribute a large-scale dataset, which consists of various types of questions and multiple sports for VideoQA.  
To our knowledge, it is the first dataset for complex reasoning over professional sports actions. 
    \item We propose a new Auto-Focus Transformer (AFT), where the attention focus is adaptive based on the question,  enabling the model to deal with questions requiring temporal information of various scales.

    \item We conduct extensive experiments on our dataset, including baseline study, evaluation of existing methods, and visualization of predictions. The results show the superiority of the proposed AFT for sports VideoQA. 
\end{itemize}

\section{Related Work}

\subsection{VideoQA Datasets}  
The development of VideoQA has been greatly facilitated by the emergence of various datasets, such as TGIF-QA \citep{jang2017tgif}, MSVD-QA, and MSRVTT-QA \citep{xu2017video}, DramaQA \citep{choi2021dramaqa}, NExT-QA \citep{xiao2021next}, NExT-OOD \citep{10107423}. TGIF-QA, for instance, offers a comprehensive set of four sub-tasks designed for temporal reasoning in general videos and is widely recognized in the VideoQA community. MSVD-QA and MSRVTT-QA are open-ended datasets constructed from existing video captions, while DramaQA focuses on understanding drama stories with hierarchical QAs and character-centered video annotation. NExT-QA, introduced for describing and explaining temporal actions, provides both multi-choice and open-ended questions through manual annotation. However, existing datasets center around general video understanding in daily scenarios or involve coarse-grained action/event reasoning. \textbf{Our contribution lies in fine-grained and professional analysis within sports scenarios. We present Sports-QA, a dataset that addresses aspects of description, chronology, causality, and counterfactual conditions for multiple sports of diverse characteristics.}

\subsection{VideoQA Methods} 
VideoQA poses a significant challenge as it necessitates models to grasp both spatial and temporal information from videos to answer questions. Various deep models have been developed, approaching this task from different perspectives \citep{fan2019heterogeneous,jiang2020reasoning,li2022invariant,antol2015vqa,gao2018motion,10172254,9770842,10214041,10146482}.
For instance, the deep heterogeneous graph alignment network by Jiang et al. \citep{jiang2020reasoning} addresses VideoQA by simultaneously aligning intra/inter-modality information. Another approach involves a multimodal attention model proposed by Fan et al. \citep{fan2019heterogeneous}, where heterogeneous memory learns global context from visual features, and question memory captures the complex semantics of questions.
IGV, introduced by Li et al. \citep{li2022invariant}, grounds question-critical scenes in videos by considering causal relations that remain invariant to complement contents.
However, a limitation in these approaches is the lack of consideration for the fact that different questions may require temporal dependencies of specific scales. \textbf{In response to this challenge, we propose the Auto-Focus Transformer, designed to automatically focus on a specific temporal scale based on the question.}

\subsection{Sports Video/Text Understanding} Sports video/text understanding has drawn increasing attention in recent years \citep{yuan2021spatio,li2021groupformer,koshkina2021contrastive,zhu2022fencenet,martin2020fine,wang2022shuttlenet,yang2024sports,wiseman2017challenges}. Researchers have made great efforts in various tasks such as sports action recognition \citep{shao2020finegym}, multi-person action detection \citep{li2021multisports}, and action quality assessment \citep{tang2020uncertainty}. Meanwhile, numerous sports datasets are constructed \citep{parmar2019action,li2021multisports,giancola2018soccernet,deliege2021soccernet}.  
AQA-7 \citep{parmar2019action} is constructed for professional action quality assessment. Seven types of actions are included in this dataset and all the action instances are associated with quality scores. \mrr{A large-scale dataset of basketball game descriptions paired with detailed statistics is introduced by \cite{wiseman2017challenges}, enabling fully data-driven modeling approaches.}
MultiSports \citep{li2021multisports} is proposed for the spatial-temporal detection of professional sports actions. FineGym~\citep{shao2020finegym} focuses on fine-grained activity localization that requires an accurate understanding of the atomic level of a gymnastic action.
\mr{Sports Intelligence \citep{yang2024sports} focuses on comprehensive LLM evaluation in sports-related NLP tasks. While it covers diverse textual scenarios, it does not address the multimodal or temporal complexities of sports video QA. In contrast, our dataset is the first to specifically benchmark LLMs in the context of professional sports video understanding, offering a new challenge for multimodal reasoning.}

\subsection{Sports VideoQA} 
\mr{SportQA \citep{xia2024sportqa} presents over 70,000 multiple-choice questions to evaluate sports knowledge in LLMs. However, it is purely text-based and lacks any video content. In contrast, our Sports-QA is a video-centric QA benchmark, targeting the unique temporal and visual reasoning challenges present in sports.
SPORTU \citep{xia2024sportu} introduces both text- and video-based QA. However, SPORTU-video consists of slow-motion clips with generic QA focused on rule application and basic event recognition. Our Sports-QA differs in three key ways: 
1) It uses high-tempo professional sports broadcasts, which are more fine-grained and realistic.
2) It introduces structured question types aligned with video understanding (e.g., descriptive, chronological, causal, and counterfactual reasoning).
3) It targets higher-level semantic reasoning grounded in temporal dynamics, not just static recognition.}

\section{Sports-QA Dataset}

\begin{table}[tbp]
\caption{The statistics for the MultiSports dataset are presented.} 
\label{mss}
\begin{tabular}{@{}lcrcc@{}}
\toprule
 Sports          & \# Action & \# Instance & Avg. Action/Video Duration & \# Bounding box \\ \midrule
Gym        & 21      & 8,703     & 1.5s / 30.7s        & 325K    \\
Volleyball & 12      & 7,645     & 0.7s / 10.5s        & 139K   \\
Football   & 15      & 12,254    & 0.7s / 22.6s        & 225K    \\
Basketball & 18      & 9,099     & 0.9s / 19.7s        & 213K   \\\midrule
Total      & 66      & 37,701    & 1.0s / 20.9s        & 902K   \\ \botrule
\end{tabular}
\end{table}

\subsection{Data Source}

Regarding the collection of sports videos, we consider the following aspects: 1) The visual quality of the video data should be high enough to conduct fine-grained video understanding, such as video resolution and frame rate. 2) Instead of applying to a single type of sports, we expect a VideoQA dataset involving multiple sports.
After a deep survey of the works on sports video understanding, we find that the MultiSports \citep{li2021multisports} and FineGym \citep{shao2020finegym} datasets are highly suitable for our purpose. The details of MultiSports and FineGym are as follows.

\noindent\textbf{MultiSports} \citep{li2021multisports} is a dataset for the temporal localization of sports actions, encompassing four sports (i.e., basketball, football, volleyball, and aerobic gymnastics) and 66 fine-grained action categories\footnote{Project homepage: \url{https://deeperaction.github.io/datasets/multisports.html}}. The action categories in MultiSports are professional terms (such as ``\textit{volleyball spike}", ``\textit{football tackle}", and ``\textit{basketball defensive rebound}") instead of common and atomic actions like ``\textit{run}" and ``\textit{stand}". For each action instance, the dataset provides the bounding boxes of a player from the starting frame of an action to the ending frame, forming the action spatial-temporal tube. The statistics of MultiSports are shown in Table \ref{mss}.

\begin{figure}[tbp]
\centering
\includegraphics[width=\columnwidth]{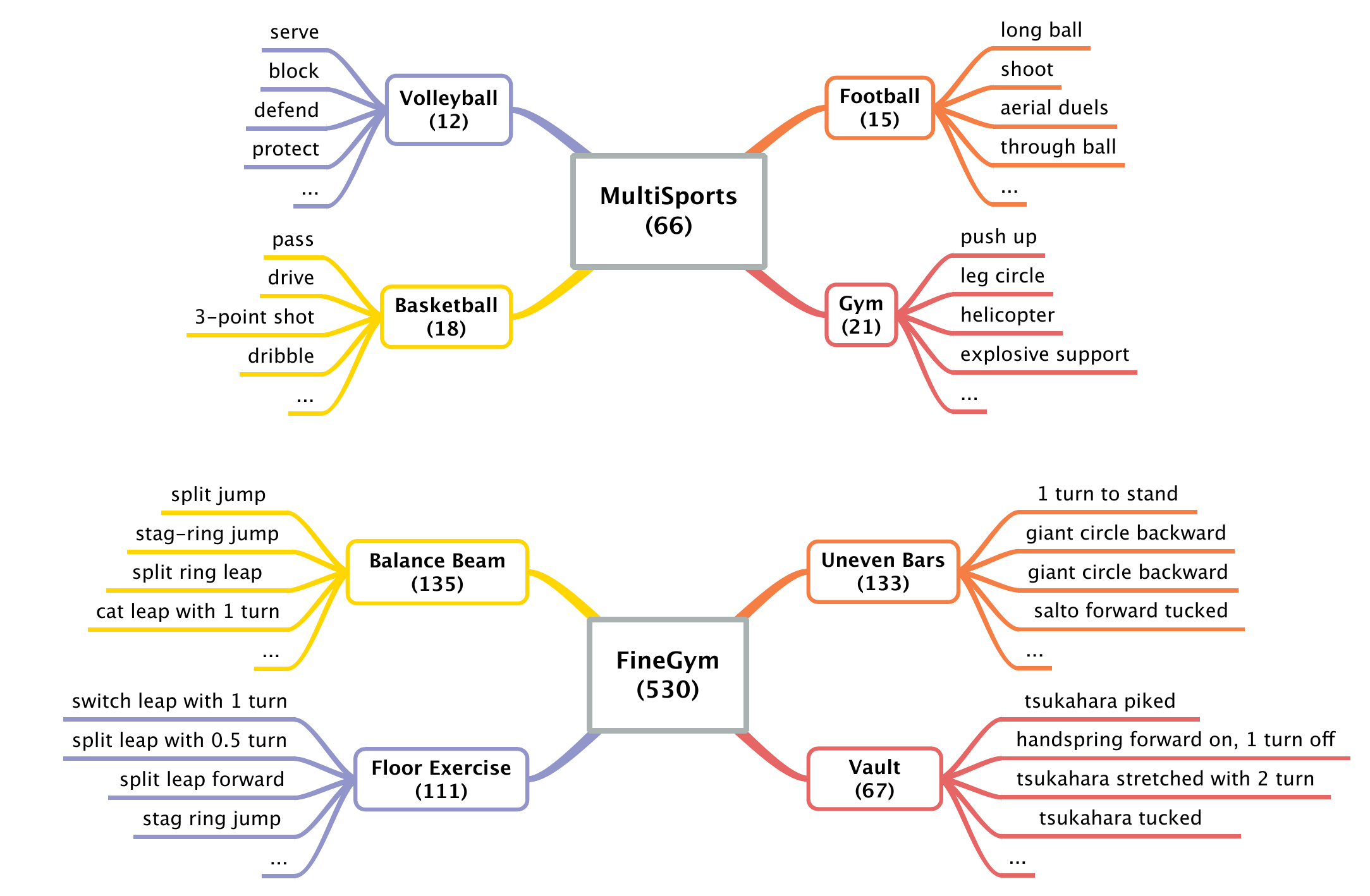}
\caption{The action hierarchy of the MultiSports dataset is depicted at the top, while the FineGym dataset's hierarchy is shown at the bottom. It's important to note that the figure includes only four example actions for each sport.}
\label{ms}
\end{figure}

\noindent\textbf{FineGym} \citep{shao2020finegym} is a dataset designed to elevate the field of action recognition by addressing the limitations observed in existing techniques\footnote{Project homepage: \url{https://sdolivia.github.io/FineGym}}. Developed to surpass current benchmarks, it offers a unique combination of richness, quality, and diversity in its content.
This dataset is constructed on gymnasium videos, providing a realistic and varied environment for action recognition studies. What sets FineGym apart from other datasets is its meticulous temporal annotation at both action and sub-action levels, featuring a three-level semantic hierarchy. This hierarchical structure allows for a more nuanced understanding of activities, enabling researchers to explore and analyze actions in finer detail. As an illustrative example, an event is annotated as a sequence of elementary sub-actions Importantly, each sub-action within these sets is further annotated with finely defined class labels, contributing to a higher level of granularity in action recognition. Fig. \ref{ms} shows the action hierarchy of the MultiSports and FineGym datasets.

The action categories in MultiSports and FineGym are professional terms (such as ``\textit{volleyball spike}", ``\textit{football tackle}", ``\textit{basketball defensive rebound}", and ``\textit{vault Salto backward tucked}") instead of common and atomic actions (such as ``\textit{run}" and ``\textit{stand}"). Such professional labels are of great value to our dataset.
Since the length of videos varies greatly (from a few seconds to a few minutes), we segment the videos into clips and generate QA pairs based on these clips.

\subsection{Question-Answer Pair Generation}

In most existing works, QA pairs are generated using two approaches:
1) Automatically generating questions and answers based on video captions using a set of predefined question templates \citep{jang2017tgif, xu2017video, yang2021just}. This method is efficient, but it may introduce obvious grammatical errors or lose crucial information during the conversion of captions into questions and answers.
2) Manually annotating questions and answers through crowdsourcing \citep{xiao2021next, yu2019activitynet, garcia2020knowit}. Although this approach produces QA pairs of high quality in terms of accuracy and expression, manual annotation is time-consuming and expensive.
In this work, we aim to ensure the quality of the textual data while considering annotation costs. We achieve this by generating QA pairs using pre-defined templates based on existing labels and newly-labeled attributes of actions. Our approach guarantees grammatically error-free QA pairs that capture crucial information in videos, thanks to the careful design of sophisticated templates. Furthermore, as ball games and gymnastics have distinct characteristics, we generate the QA pairs in different ways as described as follows.

\subsubsection{QA Pair Generation for Ball Games}

Specifically, for ball games, including basketball, football, and volleyball, we define five attributes (Team, Outcome, Cause of Outcome, Cause of Action, and Effect of Action) for actions. We then manually annotate the attributes of each action. The definitions and annotation process of these attributes are elaborated as follows.

\noindent\textbf{Attribute Definition.} MultiSports consists of 45 action categories from ball games, with some being crucial for professional sports statistics or quantifying the performance of players/teams. For instance, the ``\textit{2-Point Field Goal Percentage}" in a basketball match necessitates a fine-grained understanding of all ``\textit{2-point shot}" actions, while the ``\textit{saving}" actions in football directly reflect the performance of goalkeepers. \mr{In this work, we focus on 28 crucial actions, listed in Table \ref{acts_}.} Considering both practical applications and research purposes, we define five attributes for each of the crucial action instances as follows.

\noindent 1. \texttt{Team}: The team that the player of the action belongs to is denoted by two options: \texttt{left team} or \texttt{right team}. Specifically, in basketball and football, the determination of the left or right team depends on the team attacking towards the right or left. In volleyball, the left or right team is designated based on their position relative to the left or right side of the net.

\noindent 2. \texttt{Outcome}: This attribute assesses if the action meets the expectation and is defined as a binary label: \texttt{successful} or \texttt{failed}. For example, if a ``\textit{2-point shot}" in basketball scores a goal, it is annotated as successful; otherwise, it is labeled as failed.

\noindent 3. \texttt{Cause of Outcome}: 

This attribute indicates the cause of failure for offensive actions, such as a ``\textit{2-point shot}" in basketball, ``\textit{shoot}" in football, and ``\textit{spike}" in volleyball, or the cause of success for defensive actions, like ``\textit{block}" in football and ``\textit{save}" in volleyball. Specifically, the cause of failure for offensive actions reflects player shortcomings. For example, the failure of an offensive ``\textit{long pass}" in football could be attributed to \texttt{defensive interception}, \texttt{bad pass}, or \texttt{bad catch}. Identifying the precise cause helps teams identify their weaknesses and improve performance. Similarly, the cause of success for defensive actions highlights player strengths. For instance, the success of a defensive ``\textit{save}" in volleyball might be due to \texttt{offensive out of bounds}, \texttt{offensive blocked by net}, or \texttt{actually catch the ball}, with the last case directly reflecting defensive performance. We do not consider the cause of success for offensive actions or the cause of failure for defensive actions, as they cannot be attributed to different cases. For example, the failure of ``\textit{football saving}" can only be explained by the goalkeeper missing the ball. Therefore, these causes are not worth discussing. We define various cause options for applicable action categories and ask annotators to choose from them. Note that this attribute is applicable to certain action categories, and the causes vary for different action categories.

\noindent 4. \texttt{Cause of Action}: This attribute indicates the actions that cause the current action. Each action instance in a video is initially labeled with its time order, based on the start time, which serves as the unique ID of the action in the video. For each crucial action, annotators are asked to provide the IDs of the actions that cause the current action; these causes are not necessarily crucial actions. Consequently, the \texttt{Cause of Action} is a list consisting of the IDs of actions that are mostly temporally adjacent to the current action. Not all crucial action categories are required to provide this attribute because the causes of some actions are unique. For instance, the only cause of ``\textit{volleyball first pass}" is ``\textit{volleyball serve}".

\noindent 5. \texttt{Effect of Action}: This attribute aims to identify the actions caused by the current action, analogous to \texttt{Cause of Action}. Although the cause and effect are conjunctive, we may consider only one of them, as some actions are not crucial. For instance, if ACTION M (crucial action) causes ACTION N (non-crucial action), we only label ACTION N as an effect of ACTION M.

\noindent\textbf{Attribute Annotation.}
After defining the attributes mentioned above, we proceed to annotate them for each crucial action instance in MultiSports. The annotation process is divided into three stages: pre-annotation, formal annotation, and quality check.
In the pre-annotation stage, annotators are grouped into three categories, with each group assigned responsibility for a specific type of ball game. Each annotator within a group is then assigned several videos corresponding to their designated sport. During this stage, annotators are tasked with labeling the attributes of the first 50 action instances. Following this, an intra-group check is conducted to address any issues related to understanding bias or potential mistakes, as well as to handle unexpected or rare situations. The annotation can only progress to the next stage when annotators achieve a consistent understanding of each attribute, as outlined in the protocols.
Moving to the formal annotation stage, annotators use shared protocols to amend their previous annotations and label the remaining action instances. Once all actions are labeled, we perform an inter-group quality check on all annotations. The purpose of this inter-group check is to ensure that the annotations align with the common understanding of average individuals.
The annotation process is completed within one month and involves the collaboration of 15 graduate students.

\begin{figure*}[tbp]
\centering
\includegraphics[width=0.93\textwidth]{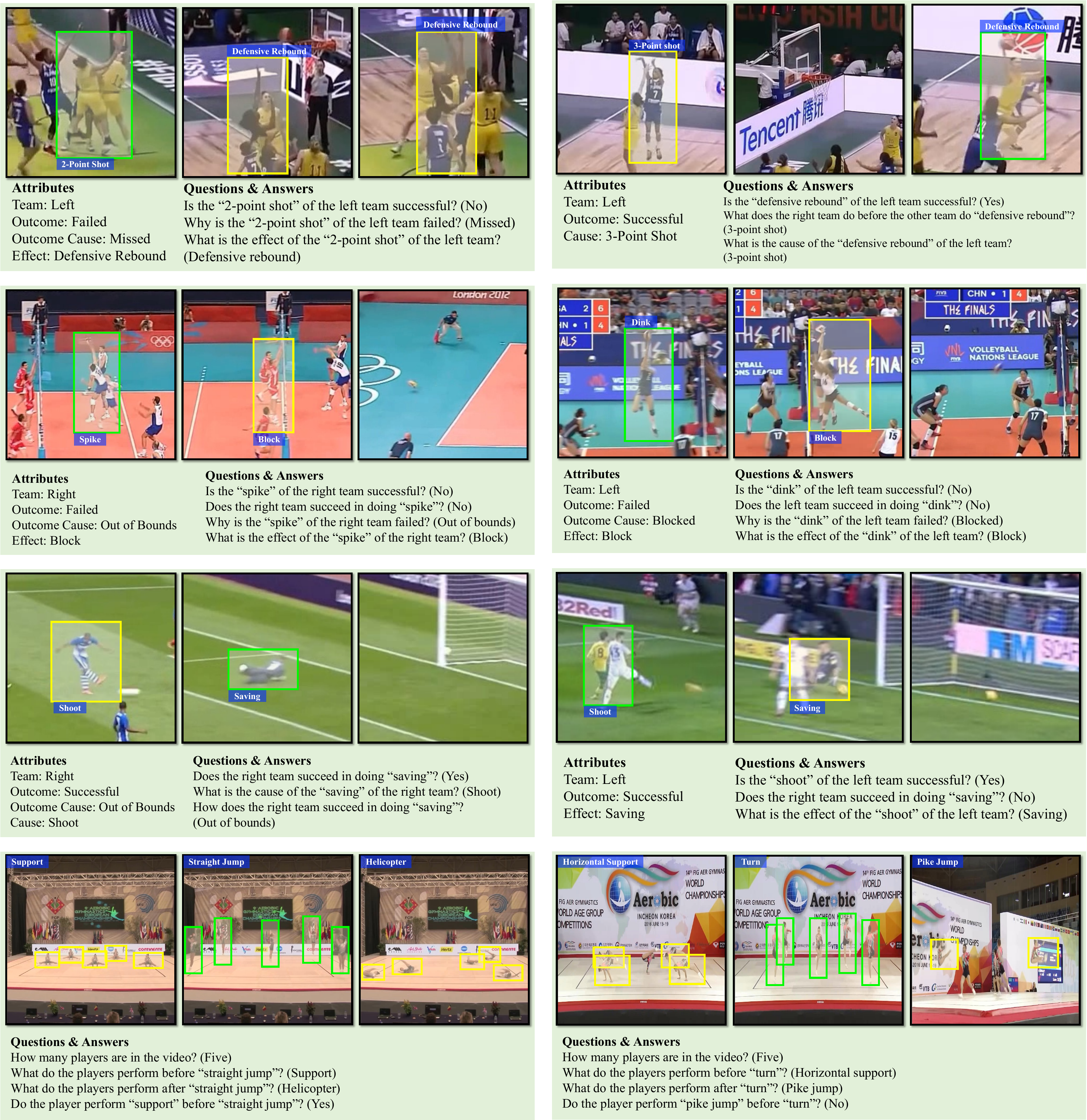}
\caption{Example of Sports-QA: The actions in the green boxes (such as ``\textit{2-point shot}") are the query actions, while the actions in the yellow boxes (such as ``\textit{block}") represent the effects. For ball games, annotators provide attribute labels, and we generate QA pairs based on these attributes. In gymnastics, we generate QA pairs using annotations from MultiSports/FineGym.
}
\label{egdata}
\end{figure*}

Based on the attributes, we generate questions about the videos by designing various templates (listed in the end of the paper) to inquire about these attributes. For instance, given a video shown in Fig. \ref{egdata}, after annotating the attributes for the action ``\textit{spike}", a question querying a particular attribute (e.g., Outcome) of the action can be generated, such as ``\textit{Is the `spike' of the right team successful?}".
Specifically, our dataset involves four types of questions: descriptive, temporal, causal, and counterfactual. The details of each type are described as follows.

\noindent 1. \textbf{Descriptive:} Descriptive questions require holistic comprehension and involve various aspects of information from the videos. These questions include simple queries such as ``\textit{What is the video about?}" and ``\textit{Does SOME-TEAM perform SOME-ACTION?}", as well as complex ones involving counting and the outcome of actions, such as ``\textit{How many times does SOME-TEAM perform SOME-ACTION?}" and ``\textit{Does SOME-TEAM successfully do their i-th SOME-ACTION?}"

\noindent 2. \textbf{Temporal:} Temporal questions focus on the temporal relations among the actions of the same or different teams. Specifically, these questions require an understanding of intra-group temporal relations and inter-group ones. Typical templates for this type of question are ``\textit{What does SOME-TEAM do before/after their i-th SOME-ACTION?}" and ``\textit{What does the left/right team do before/after the other team performs their i-th SOME-ACTION?}"

\noindent 3. \textbf{Causal:} Causal questions aim to uncover the causal relations among the action instances and to explain the reasons or the specific process of the actions. Different from temporal questions, these questions require causal reasoning based on visual cues. The templates of these questions include causal queries such as ``\textit{Why does SOME-TEAM do the i-th SOME-ACTION?}" and ``\textit{What is the effect of the i-th SOME-ACTION of SOME-TEAM?}", as well as explanation queries such as ``\textit{How does SOME-TEAM succeed in doing/fail to do the i-th SOME-ACTION?}"

\noindent 4. \textbf{Counterfactual:} Different from the above three types of questions querying the details that actually happen, counterfactual questions set hypothetical conditions that do not occur in the video and query about the expected outcomes based on the conditions. This type of question requires reasoning about various imagined situations and expects the outcomes according to the causal relations among actions. The template for this type of question is ``\textit{Would the left/right team succeed in do the i-th SOME-ACTION if the other team did not do SOME-ACTION?}"

\noindent\textbf{From Attributes to Question-Answer Pairs.} In particular, the questions are queries about the attributes, and their answers can be obtained directly from our annotations or through logical reasoning and statistical analysis over the annotations. For instance, for a question asking about the cause of a specific action, the answer can be easily retrieved from its attribute of cause; For a question asking whether a specific action is successful, we can check its outcome attribute. We guarantee the correctness of the answers as we have checked the annotations of attributes as mentioned previously.
These answers include responses like ``\textit{yes/no}", numerical values, sports names (such as ``\textit{volleyball}"), action categories in MultiSports and FineGym (such as ``\textit{block}"), and short phrases describing the reasons (such as ``\textit{out of bounds}"). See Fig. \ref{egdata} for more concrete examples of the attributes and the generated QA pairs.

An interesting aspect of our dataset is that it can be utilized for other tasks due to our two-stage annotation process. Specifically, we have defined several attributes for each action instance, which can be considered as multiple labels. Consequently, our dataset can be employed for conducting multi-label classification.
Moreover, by altering the template from question answering to declarative sentences, we can generate descriptive or explanatory narrations for sports videos. Hence, our dataset can also be exploited for sports video captioning.

\subsubsection{QA Pair Generation for Gymnastics}

A major difference between aerobic gymnastics/FineGym and ball games is the absence of the concept of a team. Furthermore, there are no causal relations among the actions performed in these sports. Therefore, we only generate descriptive and temporal questions for aerobic gymnastics/FineGym based on their annotations. In addition to the templates used for ball games, we have also designed some new templates. For descriptive questions, we include queries about the number of players or actions, such as ``\textit{How many actions does the player perform?}" and ``\textit{How many times does the player perform SOME-ACTION?}" For temporal questions, we add counting questions like ``\textit{How many times do the players do SOME-ACTION before SOME-ACTION?}" Such a QA design leverages the fine-grained features in the datasets and constitutes challenging temporal reasoning that forces the models to focus on the salient sub-actions in the whole gymnastic event.

\subsection{Diversity, Debiasing and Problem Setting}

\noindent \textbf{Diversity of Questions.} The limited diversity in generating questions using pre-designed templates may not significantly impact the significance of datasets for several reasons. First, in the context of sports, there is a high level of restriction imposed by rules and specific content, resulting in a more focused attention from the audience on particular actions and events. This inherent structure in sports content narrows down the range of potential questions, making template-based question generation a feasible approach.
Secondly, the practice of template-based question generation is widely accepted and employed in the construction of VideoQA datasets \citep{jang2017tgif, xu2017video, yang2021just}. 
The adoption of this approach in reputable datasets attests to its effectiveness and acceptance. Therefore, while the questions are template-based, their relevance and significance in the context of sports are preserved, ensuring the robustness of the dataset for analysis and evaluation.

\noindent \textbf{Debiasing.} Unlike daily scenarios, players in sports are highly restricted by rules. For instance, after the left team executes a ``\textit{volleyball serve}", the right team must follow with a ``\textit{first-pass}". This results in highly correlated actions in the videos, leading to questions and answers that are also highly correlated in our generated pairs. However, such questions are often meaningless for VideoQA, as the answers can be easily inferred through rules or common sense and should be removed.
To address this, we first obtain meta-questions by removing team information and the order of actions from the original questions. For example, the meta-question for ``\textit{What does the left team do after the other team does the second spike?}" becomes ``\textit{What does the team do after the other team does spike?}" We then examine the correlations between the meta-questions and their answers. If there is only one answer to the meta-question, we remove the corresponding QA pairs. If an answer to a meta-question occurs more frequently than other answers (frequency larger than 0.5), we randomly remove the corresponding QA pairs to balance the frequencies, ensuring they are all lower than 0.5.

\noindent \textbf{Problem Setting.} Following the approach in \citep{jang2017tgif, xu2017video}, we define an open-ended task based on the generated QA pairs. It is important to note that the open-ended setting, along with the multiple-choice setting, is widely used in the field of VideoQA. We have chosen to commit to the open-ended task instead of the multiple-choice one for two reasons: 1) The open-ended task is more challenging as it requires models to choose from a large answer set rather than selecting from several given options. 2) Limiting the answer choices to only several options would significantly decrease the diversity of the dataset. Specifically, all answers in the QA pairs form an answer pool, which is treated as 191 classes (after discarding classes with fewer than 30 samples). 

\noindent\mrr{\textbf{Interest of Different Participants.} We would like to emphasize that the questions in our dataset are intentionally designed to reflect the information needs of different participants, including audience members, commentators, referees, and coaches. For example, we include questions about the outcomes of basketball shots and football attempts, which are typically of interest to both the audience and commentators. We also design questions about the causes or intentions behind specific actions, which can assist coaches in training and instruction. In addition, we include questions related to counting technical movements in gymnastics, which are relevant to referees for scoring purposes. In summary, the question design is grounded in the real interests of these different participants.}

\noindent \textbf{Why not Fine-Grained Action Recognition?} It is essential to underscore that merely achieving fine-grained action recognition is insufficient for Sports-QA. The nature of the questions in this context demands not only precise detection of actions but also involves intricate temporal and causal modeling. To illustrate, questions that inquire about the effects of certain actions necessitate the accurate identification of temporally adjacent actions and a nuanced understanding of their causality. In summary, Sports-QA requires a more advanced level of comprehension and temporal reasoning beyond basic action recognition.

\subsection{Dataset Statistics}

Table \ref{n_qa} presents the numbers of QA pairs for various question types and different sports in Sport-QA. Our dataset comprises approximately 94K QA pairs. It's worth noting that a substantial portion of our descriptive questions involves complex action temporal localization and counting, presenting challenges similar to other question types. Additionally, we have balanced the number of questions across different sports.

\begin{table}[tbp]
\caption{The numbers of QA pairs for different types and different sports.}
\label{n_qa}

\begin{tabular}{@{}lrrrrr@{}}
\toprule
 Sports       & Descriptive & Temporal & Causal & Counterfactual & Total  \\ \midrule
Basketball & 5,629 & 22    & 785    & 278      & 6,714  \\
Football   & 6,659 & 1,355 & 1,949  & 523      & 10,486 \\
Volleyball & 6,120 & 360   & 1,942  & 685      & 9,107  \\
Gym     & 6,382 & 1,997 & 0      & 0        & 8,379  \\
Floor Exercise      &  6,046      &  11,012     & 0      & 0        &   19,418     \\
Balance Beam       &   7,477     &   12,773    & 0      & 0        &   20,250     \\
Uneven Bars      &   7,294     &   12,124    & 0      & 0        &    17,058    \\
Vault      &   2,661     &   0    & 0      & 0       &   2,661     \\ \midrule
Total   &   48,268     &   39,643    & 4,676  & 1,486    &    94,073    \\ \botrule
\end{tabular}
\end{table}

\begin{figure*}[tbp]
\includegraphics[width=\textwidth]{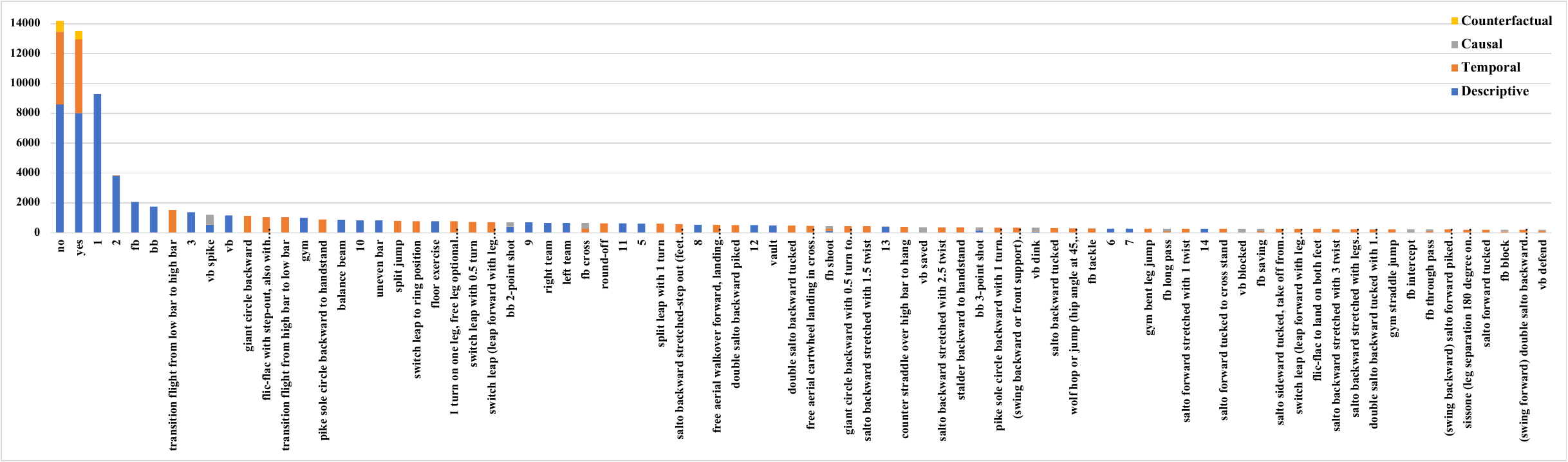}
\caption{The distributions of answer classes broken down by question types.}
\label{ans_dist_type}
\end{figure*}

The total number of videos in our dataset is 5,967, distributed across different sports as follows: basketball (879), football (1,030), volleyball (586), gym (505), vault (501), uneven bars (834), floor exercise (770), and balance beam (862). Sports-QA is divided into training, validation, and testing sets (60\%/20\%/20\%). Specifically, videos of the same sport are randomly assigned to each set to ensure similar distributions of sport types across subsets. Additionally, we have ensured that the distributions of answer classes are similar in different subsets.

Fig. \ref{ans_dist_type} illustrates the distribution of answer classes broken down by question types in Sports-QA. Note that we only showcase the first 80 classes with more examples. As depicted in the figure, the distribution of answers is long-tailed, and the unbalanced nature adds to the challenge of our dataset. Meanwhile, classes of the same type, such as ``yes/no", are balanced. This further increases the difficulty of our dataset, as the models are required to examine the video for answering the question instead of merely guessing the answer based on the question type. 

\mr{Additionally, we detail the statistics regarding the temporal and spatial characteristics of each question in Table \ref{stats1}. Specifically, we analyzed each question along two key dimensions: whether it contains spatial directional terms (e.g., ``left''/``right'') and whether it includes temporal relational terms (e.g., ``before''/``after''). As illustrated in the table, the distribution of questions is balanced with respect to both spatial and temporal attributes, indicating no inherent bias in these aspects.}

\begin{table}[tbp]
\caption{\mr{Statistics regarding the temporal and spatial characteristics in Sports-QA.}}
\label{stats1}
\begin{tabular}{@{}lrrrr@{}}
\toprule
\multicolumn{1}{c}{} & \multicolumn{1}{c}{Before} & \multicolumn{1}{c}{After} & \multicolumn{1}{c}{Other} & \multicolumn{1}{c}{Sum} \\ \midrule
Left                 & 893                        & 584                       & 7,382                     & 8,859                   \\
Right                & 146                        & 114                       & 7,832                     & 8,092                   \\
Other                & 19,147                     & 18,809                    & 39,166                    & 77,122                  \\ \midrule
Sum                  & 20,186                     & 19,507                    & 54,380                    & 94,073                  \\ \bottomrule
\end{tabular}
\end{table}

\section{Auto-Focus Transformer for Sports Video Question Answering}

\mr{A defining challenge in sports VideoQA is the need to reason over temporal dependencies of diverse scales involving multiple objects. For instance, a question asking about the number of actions requires the model to capture global dependencies in the video, whereas a question querying the effect of an action relies on short-term temporal information. The scale of temporal dependency required varies based on the question.} However, current Transformer-based or GNN-based VideoQA methods tend to focus on global dependencies, irrespective of the required scale of temporal information. To address this limitation, we propose the Auto-Focus Transformer (AFT), a new Transformer encoder featuring a novel multi-head Auto-Focus Attention (AFA) mechanism, designed specifically for sports VideoQA.

\begin{figure}[tbp]
\centering
\includegraphics[width=\columnwidth]{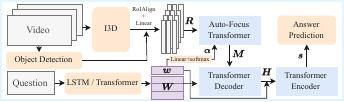}
\caption{The structure of the sports  VideoQA model based on the proposed Auto-Focus Transformer.}
\label{aft}
\end{figure}

Specifically, we start by extracting the appearance feature and motion feature of the frame using pretrained ResNet \citep{he2016deep} and I3D \citep{carreira2017quo} (refer to Section \ref{feaext} for more details). Subsequently, we concatenate these features and input them into a linear layer for dimension reduction and feature fusion. Consequently, for each video, we obtain a sequence of frame representations ($\bm{R}$). \mr{Regarding the question, we utilize a bi-directional RNN or Transformer to obtain the text representations, encompassing both the global one ($\bm{w}$) and the local ones ($\bm{W}$).}

We then  capture the dependencies among frames using AFT where the AFA mechanism
is computed as follows (the multi-head AFA is defined likewise): given the key set $\left\{\bm{k}_i\right\}$, the value set $\left\{\bm{v}_i\right\}$, and a query $\bm{q}_j$ ($\bm{q}_j,\bm{k}_i,\bm{v}_i \in \mathbb{R}^d$ are mapped from $\bm{R}$ in AFA, \mr{and $d$ is the hidden dimension}), 
\begin{equation}
{\rm AFA}(\bm{q}_j)=\sum_{f\in \mathcal{F}}\alpha_f\sum_{i\in \mathcal{D}_j^f}{\rm softmax}_i\left(\frac{\bm{q}_j^\mathrm{T}\bm{k}_i}{\sqrt{d}}\right)\bm{v}_i,
\end{equation}
where $\mathcal{F}$ is a set of pre-defined focal lengths. $\mathcal{D}_j^f=\{i||i-j|\leq f\}$ is the index set of the neighbours of $\bm{q}_j$ within focal length $f$. $\alpha_f\in(0,1)$ is the dynamic weight of focus $f$, satisfying $\sum_f\alpha_f=1$ and depended on the question. As shown in Fig. \ref{aft},   $\bm{\alpha}=\{\alpha_f\}_{f=1}^\mathcal{|F|}$ are obtained by applying linear projection\footnote{\mr{We also explored a more complex alternative structure—a two-layer MLP—and found that it yields results comparable to those of the linear projection.}} and softmax to the global representation of question.  
\mr{Since the required scale of temporal dependency is inherently question-dependent, our AFA is designed to selectively focus on only the relevant scale(s) dictated by the question. This targeted modeling ensures the model prioritizes crucial information and filters out irrelevant temporal cues, thereby enhancing the accuracy of predictions.}

We denote the output of AFT as $\bm{M}$.

For the video-text integration, we draw inspiration from \citep{li2022invariant} and \citep{jiang2020reasoning}, employing a graph convolutional network (GCN) and block fusion \citep{ben2019block}. Refer to \citep{li2022invariant} for detailed information. Ultimately, we obtain a fused feature $\bm{s}$ and predict the answer using a linear projection.

\section{Experiments}

\subsection{Experimental Settings}
\label{feaext}

We start by explaining the experimental settings, covering data pre-processing and feature extraction, implementation details, and the evaluation metrics.

\noindent\textbf{Pre-processing and Representation.} The videos in our dataset are down-sampled to 5 FPS to reduce temporal redundancy. We extract two types of visual features for experiments: appearance feature and motion feature. For each feature type, we obtain both global and local features.
Specifically, we detect 10 objects in every two frames of each video using Faster-RCNN \citep{ren2015faster} (with ResNet-101 \citep{he2016deep} as the backbone and a score threshold set to 0.6), pretrained on COCO \citep{lin2014microsoft}. These detected objects serve as local appearance features in our work. Additionally, we use the global average pooling of Res4 activation in Faster-RCNN as the global appearance feature.
To extract motion features, we employ I3D \citep{carreira2017quo} pretrained on Kinetics \citep{carreira2017quo}. Specifically, for each frame, we combine it with its adjacent 7 frames, creating a clip that we then feed to I3D to extract a 3D feature map. The global motion feature of a frame is obtained by applying global average pooling to the 3D feature. The local motion feature of an object in a frame is obtained by applying RoIAlign \citep{he2017mask} (output size $1\times 1$) to the 3D feature with its corresponding bounding box. Note that our method uses only the global feature.
For language, we explore two types of word embeddings: GloVe \citep{pennington2014glove} and pre-trained BERT \citep{devlin2018bert}.
 
\noindent\textbf{Implementation Details.} We set the attention focus to $\mathcal{F}=\{3,9,80\}$ to cover short-term, mid-term, and long-term dependencies, respectively. The hidden dimension $d$ is set to 512. The loss function is cross-entropy. The models are trained for 50 epochs with a learning rate of $1\times 10^{-4}$ and a batch size of 16, using the Adam optimizer \citep{kingma2014adam}.

\noindent\mr{\textbf{Evaluation Metrics.} For evaluation metrics, since we structure the task as a multi-class classification problem, we employ the accuracy of multi-class classification to showcase the models' performance on our dataset. Additionally, recognizing the long-tail distribution of our dataset, we evaluate methods using F1-score. F1-score is particularly important for sports VideoQA as it explicitly reflects performance averaged over imbalanced classes.}

\begin{table}[tbp]
\caption{\mr{Baseline results on Sports-QA, including random choosing, semantic-aware (S-A) random choosing, semantic-aware (S-A) most-frequent-answer choosing, LSTM-based BlindQA, and Transformer-based BlindQA. The best and the second-best results (\%) are \textbf{bold} and \underline{underlined}.}}
\label{bl}
\begin{tabular}{@{}cccccccc@{}}
\toprule
\multirow{2}{*}{Metric} & \multicolumn{3}{c}{Baseline} & \multicolumn{2}{c}{GloVe}            & \multicolumn{2}{c}{BERT}                    \\\cmidrule(l){2-4} \cmidrule(l){5-6} \cmidrule(l){7-8} 
                        & Random   & S-A Random  &S-A Frequent & LSTM & Trans.                  & LSTM                          & Trans. \\ \midrule
Accuracy                & 0.5      & 24.6  &      26.8 & 44.0   & 43.7                          & \underline{43.9} & 43.9        \\
F1-score                & 0.3      & 5.4   &    5.7   & 13.5 & \underline{14.9} & 13.8                          & 15.8        \\ \bottomrule
\end{tabular}
\end{table}

\begin{table*}[tbp]
\caption{The results of different methods on Sports-QA. The sport-wise accuracy, question-type-wise accuracy, overall accuracy and F1-score are presented (\%).
The best and second best results are \textbf{bold} and \underline{underlined}.} 
\label{res1}

\resizebox{0.95\columnwidth}{!}{\begin{tabular}{clcccccccccll}
\toprule
 \multirow{2}{*}{Word Emb.}& \multirow{2}{*}{Model} & \multicolumn{5}{c}{Sports} & \multicolumn{4}{c}{Question Type} & \multirow{2}{*}{Acc.} & \multirow{2}{*}{F1-score} \\ \cmidrule(lr){3-7}\cmidrule(lr){8-11}
 &  & Basket. & Foot. & Volley. & Gym  &FineG.& Desc. & Temp. & Causal & Counter. &  &  \\ \hline\hline
 & \cellcolor[HTML]{C0C0C0}BlindQA & \cellcolor[HTML]{C0C0C0}33.4& \cellcolor[HTML]{C0C0C0}62.9& \cellcolor[HTML]{C0C0C0}44.9& \cellcolor[HTML]{C0C0C0}36.5&\cellcolor[HTML]{C0C0C0}42.0& \cellcolor[HTML]{C0C0C0}51.7& \cellcolor[HTML]{C0C0C0}32.7& \cellcolor[HTML]{C0C0C0}48.7& \cellcolor[HTML]{C0C0C0}51.3& \cellcolor[HTML]{C0C0C0}43.7& \cellcolor[HTML]{C0C0C0}14.9\\
 & CoMem \citep{gao2018motion} & 72.8& 63.1& 69.8& 70.0&50.3& 77.0& 32.9& 50.0& 60.9& 57.1& 23.1\\
 & HME \citep{fan2019heterogeneous} & 64.8& 63.3& 68.5& 71.6&49.3& 76.5& 31.2& 50.4& 54.3& 56.0& 23.6\\
 & HGA \citep{jiang2020reasoning} & 71.3& 65.4& 72.2& 72.2&51.3& 78.1& 34.4& 52.0& 59.2& 58.3& 24.5\\
 & HQGA \citep{xiao2022video} & 72.8& 64.6& 70.5& 71.6&51.0& 77.6& 33.8& 54.0& 58.2& 57.9& \underline{25.0}\\
 & IGV \citep{li2022invariant} & 68.8& 66.0& 70.6& 72.4&52.1& 78.2& 34.7& 52.1& 59.2& \underline{58.5}& 24.8\\
 & MASN \citep{seo2021attend} & 71.2& 64.2& 70.1& 66.9&50.5& 76.4& 33.6& 50.2& 58.9& 57.0& 23.4\\ \cmidrule(l){2-13} 
 & Baseline (Ours) & 69.5& 65.4& 68.9& 71.8&52.3& 77.8& 35.0& 52.4& 59.9& 58.0& 23.7\\
\multirow{-9}{*}{GloVe} & AFT (Ours) & 73.9& 67.8& 69.5& 71.8&52.5& 78.9& 35.3& 55.1& 56.3& \textbf{59.2} {\color[HTML]{036400}(+1.2)}& \textbf{25.6} {\color[HTML]{036400}(+1.9)}\\ \hline\hline
 & \cellcolor[HTML]{C0C0C0}BlindQA & \cellcolor[HTML]{C0C0C0}34.7& \cellcolor[HTML]{C0C0C0}52.1& \cellcolor[HTML]{C0C0C0}45.9& \cellcolor[HTML]{C0C0C0}38.3&\cellcolor[HTML]{C0C0C0}43.9& \cellcolor[HTML]{C0C0C0}51.8& \cellcolor[HTML]{C0C0C0}33.3& \cellcolor[HTML]{C0C0C0}48.0& \cellcolor[HTML]{C0C0C0}53.9& \cellcolor[HTML]{C0C0C0}43.9& \cellcolor[HTML]{C0C0C0}15.8\\
 & CoMem \citep{gao2018motion} & 74.1& 64.0& 68.7& 70.7&49.3& 77.2& 32.1& 48.4& 55.6& 56.6& 26.4\\
 & HME \citep{fan2019heterogeneous} & 64.7& 63.4& 66.2& 70.6&49.3& 76.6& 30.7& 46.6& 53.6& 55.6& 22.9\\
 & HGA \citep{jiang2020reasoning} & 71.6& 66.5& 69.6& 70.7&51.3& 78.0& 33.8& 54.3& 56.6& 58.1& \underline{25.1}\\
 & HQGA \citep{xiao2022video} & 74.1& 63.5& 67.2& 70.8&50.3& 76.6& 33.6& 49.1& 57.9& 57.0& 25.0\\
 & IGV \citep{li2022invariant} & 72.0& 63.9& 70.4&72.4 &51.7& 78.0&34.2 & 51.9& 63.2& \underline{58.2}&23.8 \\
 & MASN \citep{seo2021attend} & 73.0& 63.0& 69.7& 66.6&50.1& 76.4& 32.5& 50.6& 61.2& 56.6& 24.1\\ \cmidrule(l){2-13} 
 & Baseline (Ours) & 70.7& 66.6& 69.2& 71.5&51.8& 78.0& 34.8& 50.5& 60.2& 57.9& 23.9\\
\multirow{-9}{*}{BERT} & AFT (Ours) & 72.3& 67.9& 70.4& 71.6&52.4& 78.3& 35.5& 56.8& 58.2& \textbf{59.1} {\color[HTML]{036400}(+1.2)}& \textbf{25.4} {\color[HTML]{036400}(+1.5)}\\ \botrule
\end{tabular}}
\end{table*}

\begin{table}[tbp]
\caption{The visual features used by different models. A./M./G./L. represents Appearance/Motion/Global/Local. Mem, IG, HL, and GNN stand for Memory, Invariant Grounding, Hierarchical Learning, and Graph Neural Network, respectively.} 
\label{fea}
\begin{tabular}{@{}lcccccc@{}}
\toprule
Model                           & Venue   & Insight            & G.A        & G.M.       & L.A.       & L.M.       \\ \midrule
CoMem \citep{gao2018motion}      & CVPR'18 & Mem             & \checkmark & \checkmark &            &            \\
HME \citep{fan2019heterogeneous} & CVPR'19 & Mem             & \checkmark & \checkmark &            &            \\
HGA \citep{jiang2020reasoning}   & AAAI'20 & GNN                & \checkmark & \checkmark &            &            \\
IGV \citep{li2022invariant}      & CVPR'22 & IG & \checkmark & \checkmark &            &            \\
HQGA \citep{xiao2022video}       & AAAI'22 & HL            & \checkmark & \checkmark & \checkmark &            \\
MASN \citep{seo2021attend}       & ACL'21  & GNN                & \checkmark & \checkmark & \checkmark & \checkmark \\\midrule
AFT (Ours)      &   &   AFA              & \checkmark  & \checkmark  &  & \\\botrule
\end{tabular}
\end{table}

\subsection{Baseline Study}

Several baselines are constructed to evaluate their performance on our Sports-QA dataset as follows.

\noindent\textbf{Random Choosing:} Answers are randomly selected from the 191 answer classes, resulting in an accuracy of approximately $\frac{1}{191} \approx 0.5\%$. The F1-score is obtained by running the random test for 200 times and taking the average.

\noindent\textbf{Semantic-Aware Random Choosing:} The randomly selected answer is constrained by the question type. For example, binary questions (beginning with ``\textit{Do}" or ``\textit{Is}") can only choose from ``\textit{yes}" or ``\textit{no}"; questions asking about the number of something (beginning with ``\textit{How many}") can only choose a number.

\noindent\mr{\textbf{Semantic-Aware Most-Frequent-Answer Choosing:} Like the Semantic-Aware (S-A) Random Choosing strategy, this approach operates within the constraint of question type. The key distinction lies in its selection criterion: instead of random sampling, it prioritizes the most frequently occurring answer within each question type.}

\noindent\textbf{BlindQA:} Models are constructed with a question encoder and an MLP-based answer decoder, without exploiting visual information. Two types of word embeddings (Glove and pre-trained BERT) and two question encoders (LSTM and Transformer \citep{vaswani2017attention}\footnote{\mrr{We also experimented with BERT \citep{devlin2019bert}, RoBERTa \citep{liu2019roberta}, and DeBERTa \citep{hedeberta}, as the encoder for the question encoder. However, the improvement brought by them over the baselines was marginal. Considering the balance between accuracy and model complexity, we decided not to include them in our final framework.}}) are studied.

\mr{The results of the baselines are shown in Table \ref{bl}. Random choosing exhibits poor accuracy and F1-score, but considering constraints from the questions significantly improves performance. The semantic-aware most-frequent-answer strategy yields modest improvements in both accuracy and F1-score. Importantly, however, these performance metrics remain relatively low.
This evidence strongly indicates that answer bias in our dataset is well-controlled, as even strategies explicitly leveraging frequency patterns within question types fail to achieve artificially inflated performance.
This finding reinforces the robustness of our dataset design, ensuring that model performance reflects genuine understanding rather than reliance on spurious answer distributions.
BlindQA achieves considerable results in both accuracy and F1-score, indicating noticeable semantic correlations between questions and answers. By maximizing the likelihood of the answer conditioned on a specific question type, the model fits the answer distribution given different questions in the training set.}

Regarding word embedding, pre-trained BERT achieves slightly better accuracy than GloVe, with noticeable improvements in F1-score. Concerning different question encoders, Transformer and LSTM achieve similar accuracy, but Transformer outperforms LSTM in F1-score. In summary, word embedding has a greater impact on performance than the question encoder.

\subsection{Benchmark on Sports-QA}

We benchmark several VideoQA methods on our Sports-QA dataset, comparing them to CoMem \citep{gao2018motion}, HME \citep{fan2019heterogeneous}, HGA \citep{jiang2020reasoning}, MASN \citep{seo2021attend}, HQGA \citep{xiao2022video}, and IGV \citep{li2022invariant}, with HQGA \citep{xiao2022video} and IGV \citep{li2022invariant} representing the state of the art. It's worth noting that we only evaluate methods without large-scale video-text pretraining, excluding pretrained models like MERLOT \citep{zellers2021merlot}, VIOLET \citep{fu2021violet}, and All-in-one \citep{wang2022all}. The exclusion is based on the following reasons: 1) pretrained models require large-scale additional data, making comparisons unfair for methods without pretraining; 2) we aim to highlight the characteristics of our dataset, and the universal knowledge gained by pretrained models could introduce bias in understanding sports scenarios. Table \ref{fea} shows the types of visual features used by each model, along with a summary of each method's insights. For CoMem, HME, and HGA, we use re-implementations provided by \citep{xiao2021next}, and for MASN, HQGA, and IGV, we utilize the official implementations by their respective authors. \textbf{The baseline model has the same architecture as shown in Fig. \ref{aft}, with AFT replaced with a standard Transformer Encoder.}

In Table \ref{res1}, the VideoQA methods demonstrate substantial improvements compared to BlindQA. Notably, the most significant improvements are observed for basketball (approximately 40\% increase in accuracy), indicating the critical role of visual information in basketball for question answering. Conversely, the improvements for football are less pronounced (about 4\%), suggesting that BlindQA already achieves considerable accuracy in this sport. The results suggest high correlations between questions and answers in football, posing challenges for VideoQA models to fully exploit visual information, potentially due to the complexity introduced by more players.

Examining different question types, the greatest improvements are observed for descriptive questions (around 27\% increase in accuracy). Additionally, VideoQA performance on causality and counterfactual questions shows significant improvements compared to BlindQA. However, results for temporal questions are similar between VideoQA methods and BlindQA (except for our AFT), suggesting that VideoQA models may struggle with modeling fine-grained temporal dependencies among actions.

Among the different VideoQA models, our AFT achieves the best performance in both accuracy and F1-score. The significant improvements compared to the baseline indicate the effectiveness of the proposed attention mechanism in distinguishing semantically similar classes.

\subsection{Further Analysis}

\noindent\textbf{Generalization Ability across Sports}. In this study, our primary objective is to assess the generalization capabilities across different sports domains. Our approach involves an initial pretraining phase on a set of four sports, followed by fine-tuning on a fifth, distinct sport. We meticulously evaluate the performance of this pretraining-fine-tuning methodology in comparison to training a model entirely from scratch.
The outcomes, detailed in Table \ref{resf}, showcase a notable enhancement in accuracy when employing the pretraining strategy on diverse sports. These findings suggest that the model acquires a nuanced understanding of common semantic features during the initial pretraining. Moreover, the model demonstrates its ability to effectively generalize this acquired knowledge to previously unseen sports during the fine-tuning phase.

\begin{table}[tbp]
\caption{The table presents comparisons (accuracy \%) between models trained from scratch for individual sports and those fine-tuned using data from other sports, along with the corresponding improvements (Imp.).} 
\label{resf}

\begin{tabular}{cccccc}
\toprule
Pretraining& Basketball                       & Football                         & Volleyball                       & Gym                              & FineG. \\ \midrule
\xmark    & 57.1& 55.0& 56.4& 70.2&        48.5\\
\cmark    & 58.3& 56.5& 57.7& 71.7&        50.3\\ \midrule
 Imp.& {\color[HTML]{036400}+1.2}& {\color[HTML]{036400}+1.5}& {\color[HTML]{036400}+1.3}& {\color[HTML]{036400}+1.5}&{\color[HTML]{036400}+1.8}\\\botrule
\end{tabular}
\end{table}

\begin{table}[t]
\caption{\mr{The accuracy (\%) of various focal lengths and their combinations on the validation set. Note that the combination of the same focal length implies the use of only one focal length.}}
\label{fs}

\begin{tabular}{c|ccc|ccc|c}
\toprule
   & 1 & 3 & 5 & 7 & 9 & 11 & 80   \\ \midrule
1  & \cellcolor[HTML]{C0C0C0}58.0& / & / & 58.8& 59.1& 58.8& 58.3\\
3  & / & \cellcolor[HTML]{C0C0C0}58.6& / & 59.2& \textbf{59.3}& 58.3& 58.6\\
5  & / & / & \cellcolor[HTML]{C0C0C0}58.3& 59.0& 58.7& 58.8& 58.9\\ \midrule
7  & / & / & / & \cellcolor[HTML]{C0C0C0}58.4& / & /  & 58.8\\
9  & / & / & / & / & \cellcolor[HTML]{C0C0C0}59.0& /  & 59.2\\
11 & / & / & / & / & / & \cellcolor[HTML]{C0C0C0}58.2& 58.8\\ \midrule
80 & / & / & / & / & / & /  & \cellcolor[HTML]{C0C0C0}58.2 \\ \bottomrule
\end{tabular}
\end{table}

\noindent\mr{\textbf{Impact of Focal Length.} In this experiment, we investigate the impact of different focal lengths and their combinations on the validation set. For short-term dependency, we consider lengths of 1, 3, and 5. For the mid-term, we explore lengths of 7, 9, and 11. Regarding long-term (global) dependency, we set the focal length to the video length, which is 80 seconds. The results are presented in Table \ref{fs}.
From the results, it is evident that the model with global attention performs less effectively compared to those with local attention. However, upon applying our auto-focus attention mechanism, there is a significant improvement in performance. Notably, the combination of focal lengths 3 and 9 achieves the highest accuracy. In our final model, we incorporate the global focus, resulting in the best accuracy at 59.2\%.}

\noindent\mrr{\textbf{Performance of Multi-modal Large Language Models (MLLMs).} Following the setup outlined in \citep{yang2024sports}, we evaluated the zero-shot performance of various multimodal large language models (MLLMs) on our dataset. The corresponding results from \citep{yang2024sports} are presented in Table \ref{mllm}. 
As shown in the table, the lower scores reveal that MLLMs still have substantial scope for advancement in video-based sports recognition. This outcome underscores that video-based sports understanding remains a notable challenge for MLLMs, emphasizing the intricate nature of multimodal sports comprehension tasks. Since video-based sports recognition is primarily rooted in identifying specific sports and their associated actions, MLLMs require targeted enhancements in this domain to improve performance. 
We also observe that the CoT (Chain-of-Thought) variants of MiniGPT and ChatUniVi perform worse than their original baselines. This degradation likely arises because these models were not pretrained to generate explicit reasoning steps, and introducing CoT changes the expected output distribution from concise answers to verbose reasoning chains. The additional reasoning text often introduces noise or hallucinated visual details, which can propagate errors to the final prediction. Moreover, since our benchmark assess only short final answers, the longer CoT outputs may be penalized.
To further assess the difficulty of Sports-QA, we fine-tuned an advanced multi-modal LLM, Qwen2.5-VL-3B \citep{bai2025qwen2}, on our dataset. As shown, the zero-shot model performs poorly due to the large domain gap, while fine-tuning leads to a substantial accuracy improvement, reaching 63.71\%. Despite this improvement, the fine-tuned model still leaves considerable room for further progress, highlighting the challenging nature of our benchmark.
}

\begin{table}[t]
\caption{\mrr{The accuracy (\%) of different MLLMs (zero-shot or finetuning setting) on Sports-QA.}}
\label{mllm}
\centering
\vspace{1em}
\begin{tabular}{@{}lccccc@{}}
\toprule
\textbf{Model} & \textbf{Desc.} & \textbf{Temp.} & \textbf{Causal} & \textbf{Counter.} & \textbf{Acc.} \\ \midrule
Minigpt-4-7B \citep{zhu2023minigpt}   & 36.85                & 14.10             & 23.20           & 36.51                   & 26.70            \\
Minigpt-4-7B (CoT)   & 36.38                & 14.15             & 19.79           & 34.54                   & 26.26            \\
Chat-UniVi-7B \citep{jin2024chat}  & 45.31                & 16.23             & 24.23           & 6.58                    & 31.52            \\
Chat-UniVi-7B (CoT)  & 45.04                & 14.71             & 23.09           & 13.82                   & 30.81            \\
Video-LLaVA-7B \citep{zhu2023languagebind} & 42.80                & 13.52             & 36.80           & 42.76                   & 30.33            \\
Video-LLaVA-7B (CoT) & 43.46                & 13.50             & 35.57           & 39.47                   & 30.55            \\ 
LLaVA-Video-Qwen2-7B \citep{zhang2024video} & 45.77& 15.94& 37.73&41.78&32.73\\
Qwen2.5-7B \citep{bai2025qwen2}& 46.57&16.43& 38.41&42.87&33.40\\
Qwen2.5-3B  & 37.81& 15.45& 25.61& 37.57&27.77 \\
Qwen2.5-3B-FT & 81.71 &42.63& 58.11 &58.97 &63.71\\
\bottomrule
\end{tabular}
\end{table}

\subsection{Visualization}

\begin{figure}[tbp]
\centering
\includegraphics[width=0.8\columnwidth]{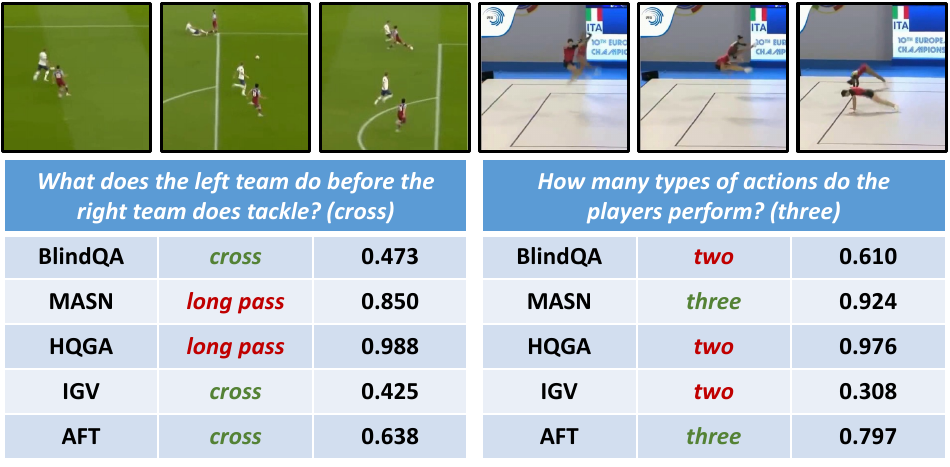}
\caption{
\mr{Visualization of the predictions. The correct/wrong predictions are highlighted in \textcolor[RGB]{84,130,53}{green}/\textcolor[RGB]{192,0,0}{red}. The predicted probability of each answer is also reported. 
}}
\label{pred}
\end{figure}

\noindent \textbf{Qualitative Results.} Fig. \ref{pred} illustrates predictions from various methods. In some instances, BlindQA successfully guesses the correct answer by leveraging the correlation between the question and the answer in the training set. However, this approach is not foolproof, as the blinded model consistently selects the most correlated answer to the question.
In comparison, the VideoQA models struggle to discern subtle differences between semantically or visually similar answer classes. For example, actions like ``\textit{long pass}" and ``\textit{cross}" both involve passing the ball to other players but differ in passing direction and distance, presenting challenges for accurate distinction. Additionally, the compared VideoQA models exhibit limitations in fine-grained motion analysis, such as action counting, as evident in the last example. With its ability to capture various types of temporal dependencies, our AFT outperforms the compared models on these challenging samples.

\begin{figure}[tbp]
\centering
\includegraphics[width=0.9\columnwidth]{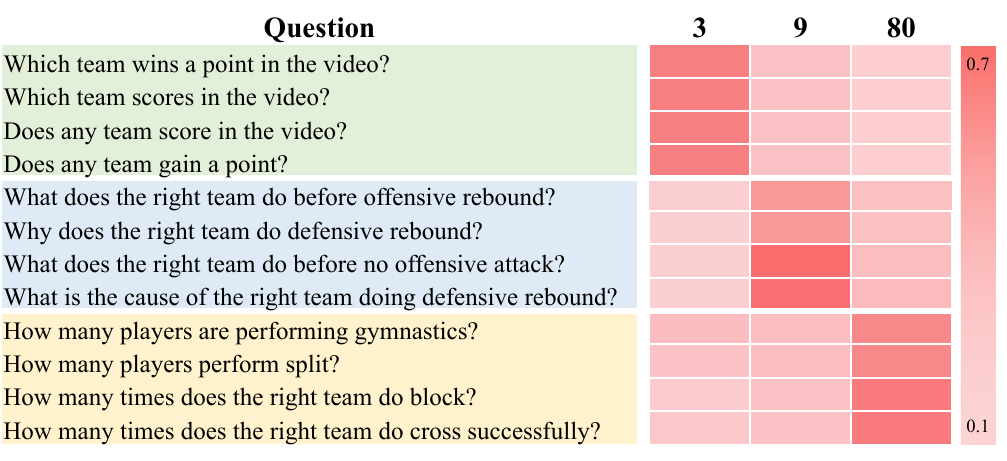}
\caption{\mr{Visualization of focus weight. We show four examples of the highest weight for each focus.}}
\label{attn_vis}
\end{figure}

\noindent\mr{\textbf{Focus Weight.}
Fig. \ref{attn_vis} depicts the focus weights of selected questions. Notably, questions seeking information about specific actions (e.g., ``\textit{win a point}") heavily rely on short-term information (foccal length 3). In contrast, questions like ``\textit{How many times does the right team do `block'?}" necessitate a stronger emphasis on long-term information. Questions pertaining to the relationships between two actions show a preference for mid-term information, offering insights into the cause and effect of actions. Intuitively, the mid-term focal length can provide more nuanced information on the interactions between actions. This visualization affirms the effectiveness of our model, which dynamically integrates different focus scales based on the diverse temporal dependencies required by various questions.}

\section{Conclusion}

We introduce Sports-QA, a novel dataset emphasizing professional action understanding, interactions among multiple players, and fine-grained motion analysis for sports VideoQA. Featuring multiple sports and diverse question types, the dataset offers a comprehensive evaluation platform for VideoQA methods. Additionally, we propose Auto-Focus Transformer (AFT), designed to capture various temporal dependencies. Our extensive experiments on the dataset demonstrate the effectiveness of AFT.

\section{Limitation and Future Work}
\label{limit}

\mr{Our proposed model does exhibit zero-shot generalization ability, precisely because it is not pretrained on large-scale multimodal datasets—this absence of prior exposure to such data is what underscores its capacity to generalize to unseen scenarios without explicit training. That said, we recognize the value in further exploring the nuances of this generalization capability, and we plan to delve deeper into this aspect as part of our future work to provide more comprehensive insights. 
\mrr{Additionally, we convert answers into 191 categories in our dataset, which limits free-form generation and semantic diversity. This formulation merely serves as a practical intermediate step between rigid classification and fully free-form generation.
We plan to explore evaluation methods that capture open-ended responses, such as semantic similarity metrics or hybrid protocols combining categorical and free-form assessments.}}

\section{Ethical Consideration}
\label{ethical}

In developing our Sports-QA dataset, sourced from MultiSports and FineGym, we prioritized ethical considerations, adhering to data privacy and usage standards while ensuring compliance with the terms of the original datasets. To address potential biases, we actively promoted diversity in sports categories and specific sports actions, aiming for a balanced representation.
Mindful of the environmental impact, we optimized our data processing to minimize resource consumption. Acknowledging the potential for misuse, we emphasize the dataset's intended use in positive sports analytics applications. Our commitment to transparency and reproducibility is evident in our methodology and dataset accessibility, ensuring that our contribution to sports analytics and computer vision is both responsible and beneficial.

\section*{List of Crucial Actions}

The selected crucial actions are listed in Table \ref{acts_}.

\begin{table}[tbp]
\caption{Crucial actions selected for Sports-QA, which are classified into offensive actions and defensive actions.} 
\label{acts_}

\begin{tabular}{@{}lll@{}}
\toprule
\multirow{2}{*}{Sports}     & \multicolumn{2}{c}{Action}                                \\ \cmidrule(l){2-3} 
                            & Offensive                    & Defensive                  \\ \midrule
\multirow{5}{*}{Volleyball} & \textit{serve}               & \textit{first pass}        \\
                            & \textit{dink}                & \textit{defend}            \\
                            & \textit{spike}               & \textit{protect}           \\
                            & \textit{second attack}       & \textit{adjust}            \\
                            & \textit{no offensive attack} & \textit{save}              \\ \midrule
\multirow{5}{*}{Football}   & \textit{shoot}               & \textit{diving}            \\
                            & \textit{cross}               & \textit{tackle}            \\
                            & \textit{through pass}        & \textit{steal}             \\
                            & \textit{long pass}           & \textit{block}             \\
                            & \textit{aerial duels}        & \textit{}                  \\ \midrule
\multirow{5}{*}{Basketball} & \textit{jump ball}           & \textit{save}              \\
                            & \textit{free throw}          & \textit{pass steal}        \\
                            & \textit{2-point shot}        & \textit{dribble steal}     \\
                            & \textit{3-point shot}        & \textit{defensive rebound} \\
                            & \textit{offensive rebound}   & \textit{}                  \\ \botrule
\end{tabular}
\end{table}


\section*{Question Templates for Sports-QA}

\subsection*{\centering Gymnastics}

\subsubsection*{Descriptive Question}
\begin{itemize}
    \item \textit{What is the video about?}
 \item \textit{How many SOME-ACTION do the players perform?}
 \item \textit{How many types of SOME-ACTION do the players perform?}
 \item \textit{How many times do the players perform SOME-ACTION?}
 \item \textit{How many players perform the i-th SOME-ACTION?}
 \item \textit{Do the players perform SOME-ACTION?}
\end{itemize}

\subsubsection*{Temporal Question}
\begin{itemize}
 \item \textit{Do the players perform SOME-ACTION before/after the i-th SOME-ACTION?}
 \item \textit{What do the players perform before/while/after performing the i-th SOME-ACTION?}
 \item \textit{How many times do the player do SOME-ACTION before/after doing the i-th SOME-ACTION?}
\end{itemize}

\subsection*{\centering Ball Games}
\subsubsection*{Descriptive Question}

\begin{itemize}
    \item \textit{What is the video about?}
\item \textit{Does one team score in the video?}
\item \textit{Which team scores in the video?}
\item \textit{How many times does SOME-TEAM do SOME-ACTION?}
\item \textit{How many times does SOME-TEAM do SOME-ACTION successfully?}
\item \textit{Does SOME-TEAM do their i-th SOME-ACTION successfully?}
\end{itemize}

\subsubsection*{Temporal Question}

\begin{itemize}
    \item \textit{What does SOME-TEAM do before/after their i-th SOME-ACTION?}
    \item \textit{What does SOME-TEAM do before/after the other team does their i-th SOME-ACTION?}
\end{itemize}

\subsubsection*{Causal Question}

\begin{itemize}
    \item \textit{Why does SOME-TEAM do the i-th SOME-ACTION?}
    \item \textit{What is the effect of the i-th SOME-ACTION of SOME-TEAM?}
    \item \textit{How does SOME-TEAM succeed in doing the i-th something?}
\end{itemize}

\subsubsection*{Counterfactual Question}

\begin{itemize}
    \item \textit{Would SOME-TEAM succeed in doing the i-th SOME-ACTION if the other team did not do SOME-ACTION?}
\end{itemize}

\section*{Acknowledgments}

This research was particularly supported by the Australian Government through the Australian Research Council's DECRA funding scheme (Grant No.: DE250100030), Discovery Project funding scheme (Grant No.:  DP260100218, DP260101891), the National Natural Science Foundation of China (No. 62372491), and the Guangdong Basic and Applied Basic Research Foundation (2022B1515020103).

\section*{Declarations}

\subsection*{Data and Code Availability}
The proposed dataset, Sports-QA, and the code of the proposed method are available at \url{https://github.com/HopLee6/Sports-QA}.

\bibliography{sn-bibliography}

\end{document}